\documentclass[letterpaper, 10 pt, conference]{ieeeconf}  

\IEEEoverridecommandlockouts                              

\usepackage{siunitx}
\usepackage{hyperref}
\usepackage[T1]{fontenc}
\usepackage{graphicx}  
\usepackage[]{subfig}
\usepackage{amsmath}
\usepackage{amsfonts} 

\usepackage{algorithm}
\usepackage[noend]{algpseudocode}
\makeatletter
\def\BState{\State\hskip-\ALG@thistlm}
\makeatother
\algnewcommand\algorithmicforeach{\textbf{for each}}
\algdef{S}[FOR]{ForEach}[1]{\algorithmicforeach\ #1\ \algorithmicdo}

\title{\LARGE \bf
Fast and Robust Detection of Fallen People from a Mobile Robot 
}

\author{Morris Antonello, Marco Carraro, Marco Pierobon and Emanuele Menegatti
\thanks{The authors are with the Intelligent Autonomous Systems Laboratory (IAS-Lab), Department of Information Engineering (DEI), University of Padova, Via Ognissanti 72, 35129, Padova, Italy. {\tt\small {morris.antonello, marco.carraro, emg}@dei.unipd.it}}
}

\begin{document}

\maketitle
\thispagestyle{empty}
\pagestyle{empty}

\begin{abstract}

This paper deals with the problem of detecting fallen people lying on the floor by means of a mobile robot equipped with a 3D depth sensor. 
In the proposed algorithm, inspired by semantic segmentation techniques, the 3D scene is over-segmented into small patches. 
Fallen people are then detected by means of two SVM classifiers: the first one labels each patch, while the second one captures the spatial relations between them. This novel approach showed to be robust and fast. Indeed, thanks to the use of small patches, fallen people in real cluttered scenes with objects side by side are correctly detected. Moreover, the algorithm can be executed on a mobile robot fitted with a standard laptop making it possible to exploit the 2D environmental map built by the robot and the multiple points of view obtained during the robot navigation. Additionally, this algorithm is robust to illumination changes since it does not rely on RGB data but on depth data. All the methods have been thoroughly validated on the IASLAB-RGBD Fallen Person Dataset, which is published online as a further contribution. It consists of several static and dynamic sequences with 15 different people and 2 different environments. 

\end{abstract}

\section{INTRODUCTION}

In the richest countries, the population pyramid is turning upside down \cite{he2016aging}. In 2015, 8.5 percent of the world's population was aged 65 and over and, by 2050, this older population is projected to represent 16.7 percent of the world total population. To allow people to continue to have active and productive lives as they age, new technologies are being studied. Recently, as far as home robots are concerned, there has been many promising developments. New products like Softbank's Pepper have been introduced into the market and many research platforms, e.g. the healthcare robots Pearl \cite{pollack2002pearl}, ASTRO \cite{cavallo2013design}, Max \cite{gross2015robot}, Hobbit \cite{fischinger2016hobbit} or our prototype O-Robot \cite{carraro2015open}, have been proposed. Not only such robots aim at fostering research to keep the house safe by monitoring and detecting anomalies, but also at being friendly companions able to enhance the elderly people's social lives without invading their privacy. In particular, among all the sources of harm, falls are known to be the major one in elderly people \cite{lord2007falls}. In this work, given that it is unlikely for a robot to capture the act of falling while patrolling, the focus is on detecting people already lying on the floor. 

The main contributions in this paper are:
\begin{itemize}
	\item a real-time pure-3D approach to detect fallen people suitable for real cluttered scenes;
	\item its integration with two basic robot functionalities, 2D mapping and navigation, in order to suppress false positives thanks to the a-priori knowledge of the environment and the availability of multiple view points; 
	\item our RGB-D dataset of fallen people\footnote{\url{http://robotics.dei.unipd.it/117-fall}} consisting of several static and dynamic sequences with 15 different people acquired in 2 different environments.
\end{itemize}

The remainder of the paper is organized as follows. Section \ref{related_work} reviews the work related to fall detection, people detection and body pose estimation. Section \ref{approach} describes our novel approach, first giving a picture of the entire workflow, then focusing on both the single-view approach and its integration with mapping and robot navigation. In Section \ref{results}, our dataset is described and our methods thoroughly evaluated. Finally, in Section \ref{conclusions}, conclusions are drawn and future directions of research identified.
\begin{figure*}
  \centering
  \includegraphics[width=\textwidth]{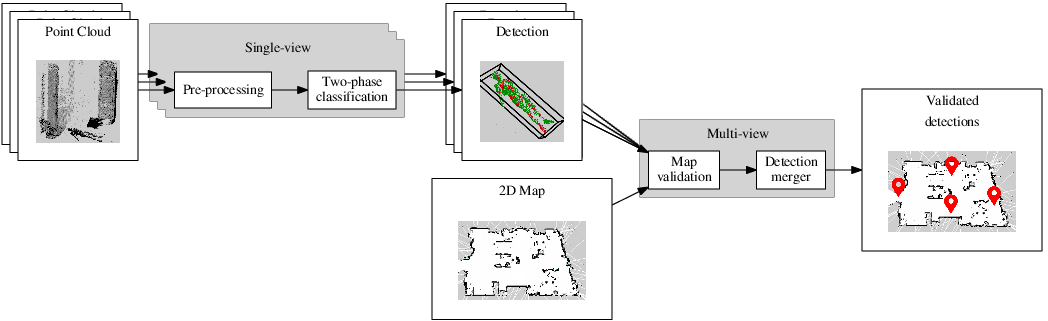}
  \caption{The proposed approach is split into two separately running processes. The \textit{single-view detector} detects fallen people on the single frames in a way which proves to be fast and robust to clutter. The \textit{multi-view analyser} fuses the single-view results exploiting the availability of the 2D map and the multiple points of view explorable during the robot navigation. The final map includes also the semantic information about the location of the fallen people, see the red placeholders.}
  \label{fig:overview}
\end{figure*}
%




\section{RELATED WORK}
\label{related_work}

Nowadays the wide adoption of Deep Neural Networks (DNN) is boosting the classification accuracy in many fields. In particular, many recent works~\cite{angelova2015real,cao2016realtime,wei2016cpm} address the person detection and body pose estimation problems showing great results. This kind of algorithms could be used also for detecting people lying on the floor. Nevertheless, their recognition capabilities are limited to RGB images and so they cannot work in dimmer scenes, which are usual in real life houses. In addition, the high complexity of the DNN requires the algorithm to be accelerated by using high-end Graphical Processing Units (GPU) in order to achieve real-time performances useful in real applications. For these reasons, those networks do not fit our application. Indeed, we are proposing techniques which can work also without the presence of the color information (e.g. under different illumination conditions or during the night). Moreover, we want to keep the power requirements at a minimum, given that this is a major issue in the design of mobile robots. Thus, the usage of an high-end GPU is unsuitable. Our approach draws upon two recent methods for the semantic segmentation of scene structures and objects from RGB-D data \cite{wolf2015fast, wolf2016enhancing}. Both approaches are almost real-time and based on fast features calculated on 3D patches or clusters. They also try to learn contextual relations among them, respectively by means of Conditional Random Fields and 3D Entangled Forests.

There exist also more specific approaches addressing the detection of falls. These comprehend wearable devices, whose great popularity is linked to the spread of open-source platforms which are small, powerful and connectible to low-cost sensors \cite{perry2009survey}. In most cases, such sensors include accelerometers \cite{li2009accurate, boyle2008simulated, lindemann2005evaluation}. These technologies suffer from the difficulty of correctly 	distinguishing falls from common actions like sitting or lying down. Furthermore, the elderlies easily forget to wear them. Other approaches specifically addressing falls need the installation of environmental devices like microphones \cite{popescu2008acoustic}, cameras for person tracking \cite{williams2007aging,cucchiara2007multi,ghidoni2014distributed}, infrared or vibration sensors \cite{yazar2014multi}. Anyway, these approaches are less effective and, being invasive, less accepted.   

To the best of our knowledge, there exist just a few approaches trying to detect fallen people already lying on the floor: \cite{wang2012lying, volkhardt2013fallen, nishi2015head}. Both \cite{wang2012lying} and \cite{volkhardt2013fallen} are specifically designed for mobile robots. In \cite{wang2012lying}, the authors propose a pipeline working on just single RGB images extending a deformable part-based model to the multi-view case for viewpoint invariant lying posture detection. Like us, \cite{volkhardt2013fallen} proposes a pipeline working on single depth images. Putative candidates are found by means of a segmentation phase based on an Euclidean clustering. Then, they are layered so as to face with occlusions and classified by means of a SVM using Histograms of Local Surface Normals. The downside of the approach is the Euclidean segmentation, in particular its distance threshold: if people fall on or near furniture, the segmented object may contain the user and parts of the furniture. On the contrary, this work specifically addresses this problem by concatenating two classifiers. Unfortunately, neither the code or dataset of \cite{volkhardt2013fallen} are available making a direct comparison impossible. Finally, in \cite{nishi2015head}, a method for detecting and locating the head of a person lying on the floor by means of a RGB-D sensor is proposed. It would allow to test vital signs on the fallen people, but has not been tested in real cluttered scenarios and requires the head to be visible. Remarkably, none of the previous approaches take advantage of the other functionalities available thanks to the mobile robot like 2D mapping, i.e. the actual knowledge of the environment, and navigation, i.e. the availability of multiple view points.


\section{APPROACH}
\label{approach}

An overview of the proposed approach for detecting people lying on the floor is given in Figure \ref{fig:overview}. It is decoupled into two separately running processes, the \textit{single-view detector} and the \textit{multi-view analyser}. The former process, the \textit{single-view detector}, operates on pure-3D Point Clouds generated by a RGB-D sensor such as the Kinect One V2, which, in our experiments, is mounted on a mobile robot 1.16\,m off the floor and parallel to it. First, the input cloud is preprocessed in order to restrict the subsequent phases to work on a region of interest comprehending all the objects above the floor and below a maximum height. Then, the pre-processed cloud is over-segmented into small patches of voxels with similar appearance. In a two-phase classification step, the patches are classified as part of person or not and gathered together. The use of the Euclidean clustering on the cloud including only the person patches makes it possible to handle also cluttered scenes. Finally, to further improve performances, the latter process, the \textit{multi-view analyser}, rejects all the detections not belonging to the free space of the 2D map and accumulates the detections from several frames by taking into account their 2D map positions and timestamps. Each phase is deeply discussed in the next subsections: Subsection \ref{subsec:patchbaseddetectionoffallenpeople} deals with the description of the \textit{single-view detector} while Subsection \ref{subsec:map_verification} and Subsection \ref{subsec:multiplevantagepoints} describe the \textit{multi-view analyser}.

\subsection{Patch-based Detection of Fallen People}
\label{subsec:patchbaseddetectionoffallenpeople}

Each point cloud is pre-processed to restrict the analysis to a region of interest and reduce the data noise. First of all, the point cloud is truncated to a 3D region containing the floor and the points between it and a maximum height of 0.66\,m. Then, the floor is removed with an approach based on the RANSAC segmentation \cite{fischler1981random}. To improve its robustness to the robot motion, two floor planes are estimated, on a first half of the cloud close to the robot and on a second half far from the robot. In particular, a good split distance proved to be 3\,m. Finally, to reduce the data noise without affecting the running time, a soft statistical outlier removal is applied with the number of neighbours set to 50 and the standard deviation set to 0.3. 

The core of the algorithm draws upon two recent works about the semantic segmentation of objects and scene structures \cite{wolf2015fast,wolf2016enhancing}. It comprehends the following 4 phases:
\begin{enumerate}
\item supervoxel over-segmentation in 3D patches;
\item classification of each 3D patch as positive, i.e. part of a fallen person or negative, i.e. not part of a fallen person;
\item clustering of positive patches;
\item classification of each cluster as positive, i.e. a fallen person, or negative, i.e. not a fallen person. 
\end{enumerate}
They allow to segment and classify correctly also the people lying close to other objects or scene structures. In the following, each phase is described.

\begin{figure}[!ht]
	\centering
	 {\includegraphics[width=\columnwidth]{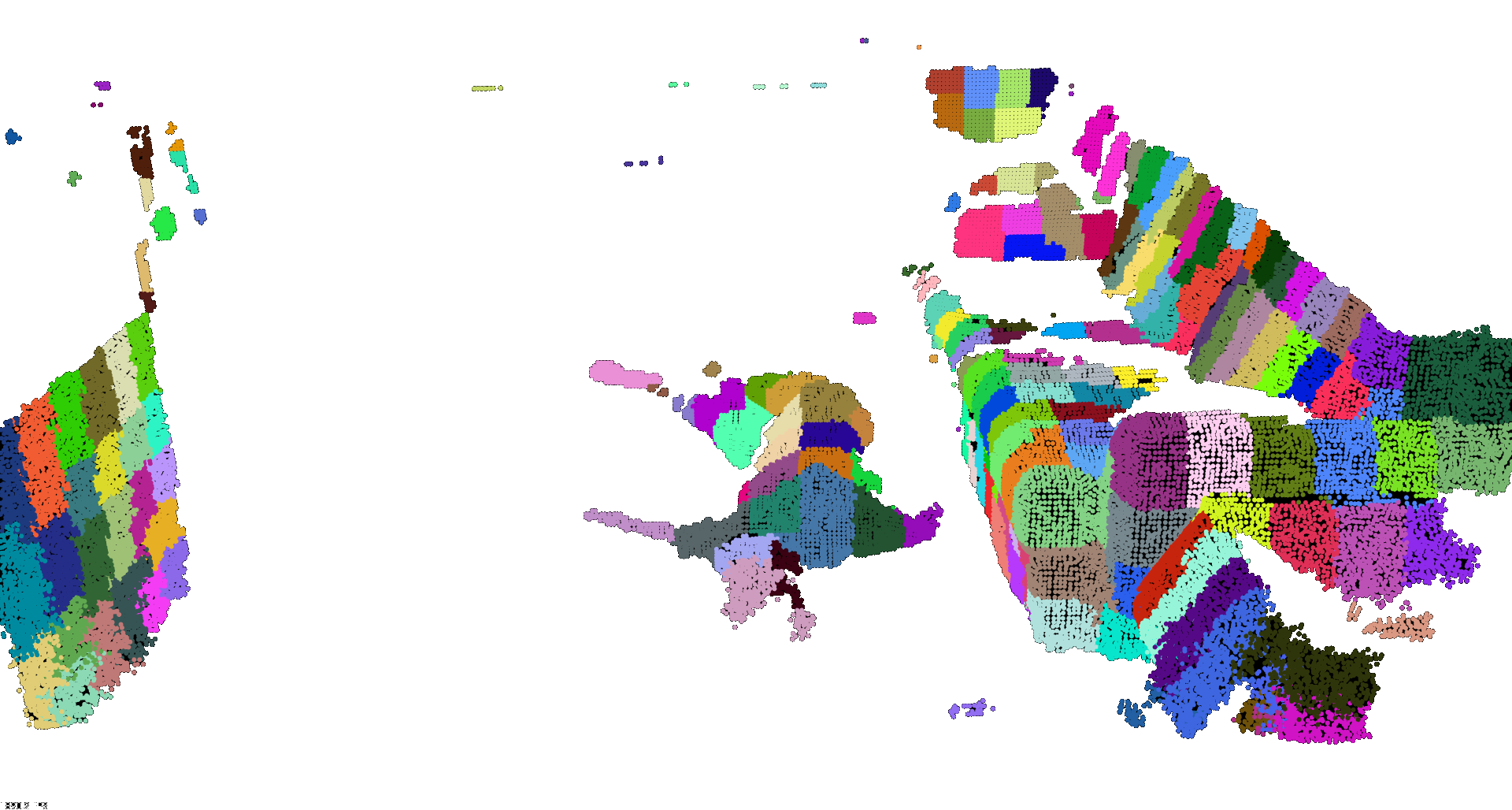}}
	 \caption{An example of pre-filtered and over-segmented cloud. A random color is assigned to each patch. The person is lying in the center.}	
	\label{fig:supervoxelsegmentation}
\end{figure}

The pre-processed point cloud is over-segmented into homogeneous 3D patches by means of the Voxel Cloud Connectivity Segmentation (VCCS) \cite{papon2013voxel}. An example of over-segmented cloud is reported in Figure \ref{fig:supervoxelsegmentation}. This solution preserves the edges by finding patches not crossing object boundaries and, at the same time, it reduces the noise and the amount of data. The set of parameters used here is: voxel resolution 0.06\,m, seed resolution 0.12\,m, color importance 0.0, spatial importance 1.0 and normal importance 4.0. The voxel resolution is a good trade off between speed and having a sufficient number of points per patch. The seed resolution is a good trade off between having big patches and over-segmenting also the thinner body elements, e.g. arms and legs. The others are suggested in \cite{stein2014convexity}. As the proposed approach does not rely on RGB data, color is not considered at all by setting the color importance to 0.0.

\begin{figure}[!ht]
	\centering
	\subfloat[]
	{\includegraphics[width=0.33\columnwidth]{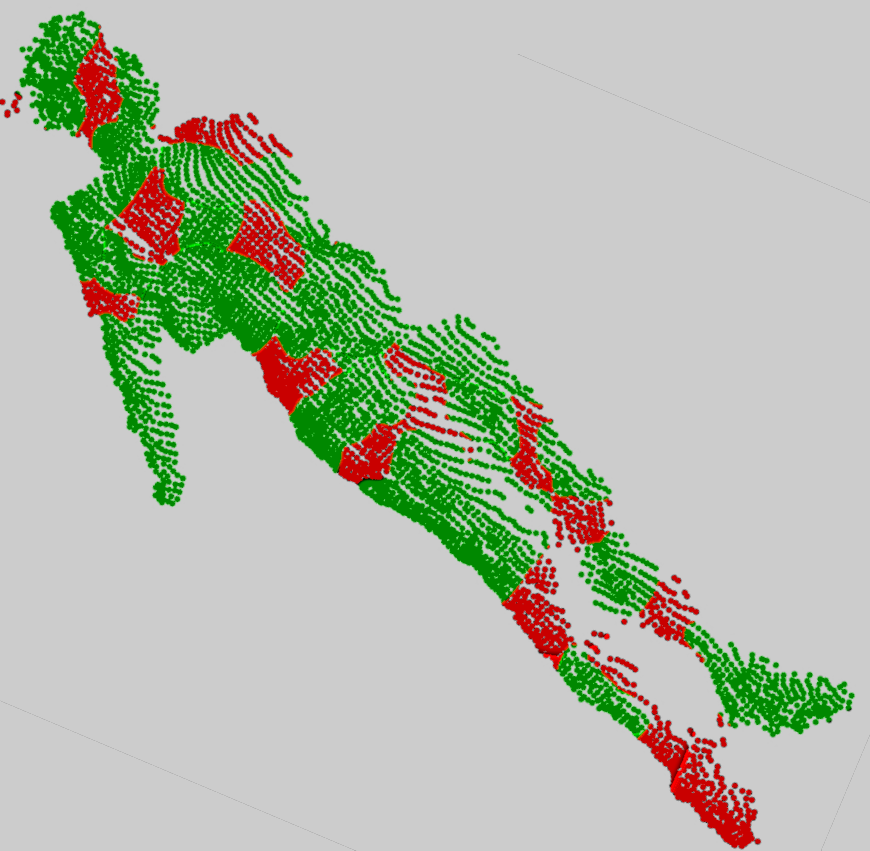}}
	\subfloat[]
	{\includegraphics[width=0.33\columnwidth]{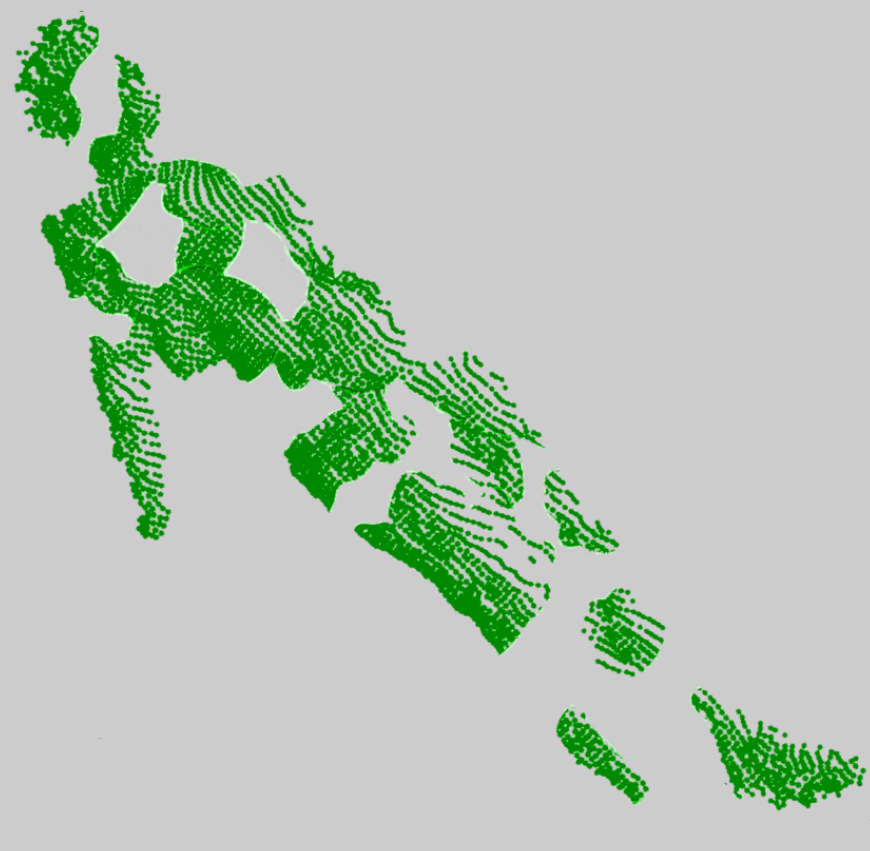}}
	\subfloat[] 
	{\includegraphics[width=0.33\columnwidth]{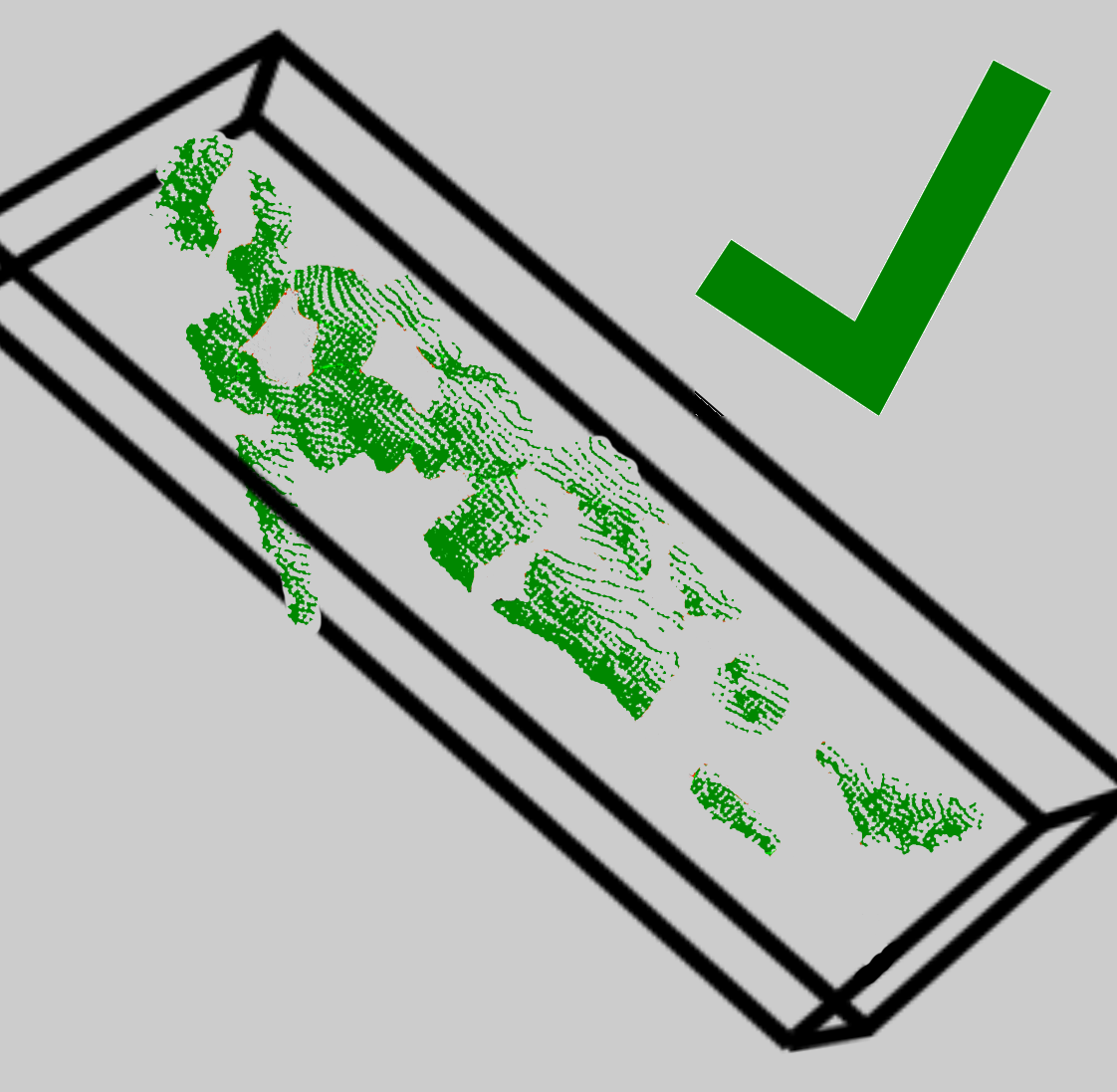}}	
	\caption{The last three steps of the algorithm core: a) The first SVM classifies each patch as a person part (green color) or not (red color); b) Euclidean clustering of the positive patches; c) Calculation of the cluster OBB. The second SVM classifies each cluster as a person or not. Here, the response is positive.}
	\label{fig:corerecap}
\end{figure}

For each patch generated by the over-segmentation, a feature vector $\mathbf{x_1}$ of length 16 is calculated. The choice of the features is based on the semantic segmentation works \cite{wolf2015fast,wolf2016enhancing}, whose presented features proved to be as fast as effective. Here, the color features are left out and only the geometric features are taken into account. Some of them are calculated from the eigenvalues of the scatter matrix of the patch, $\lambda_0 \leq \lambda_1 \leq \lambda_2$ while others from the Oriented Bounding Box (OBB) including all the patch points. The complete list is given in Table \ref{features1}.
\begin{table}[h]
\caption{List of features calculated for each 3D patch and their dimensionality.}
\label{features1}
\begin{center}
\begin{tabular}{c c}
\hline
Features & Dimensionality\\
\hline
Compactness ($\lambda_0$) & 1\\
Planarity ($\lambda_1 - \lambda_0$) & 1\\
Linearity ($\lambda_2 - \lambda_1$) & 1\\
Angle with floor plane (mean and std. dev.) & 2\\
Height (top, centroid, and bottom point) & 2\\
OBB dimensions (width, height and depth) & 3\\
OBB face areas (frontal, lateral and upper) & 3\\
OBB elongations ($\frac{height}{width}$, $\frac{depth}{width}$, $\frac{height}{depth}$) & 3\\
\hline
Total number of features & 16\\
\hline
\end{tabular}
\end{center}
\end{table}
To calculate the predicted label (part or not part of a fallen person) for each patch, this feature vector is then passed to a binary SVM classifier. After k-fold validation, a Radial Basis Function (RBF) kernel with the misclassification cost $C$ equal to 62.5 and the bandwidth $\gamma$ equal to 0.51 turned out to be the best performing solution. Of course, having each patch classified as part of a person (positive) or not (negative) does not suffice to detect a fallen person. Indeed, as shown in Figure \ref{fig:corerecap}(a), given that this classifier analyses just small patches, there can be false positives and false negatives. Because of this, two further steps, explained in the following and sketched in Figure \ref{fig:corerecap}(b)(c), have been developed in order to find 3D regions with a high density of positive patches and whose size is comparable to that of a person.

In contrast to the methods in \cite{volkhardt2013fallen}, having two sets of patches respectively with the positive and negative ones opens up the possibility to apply the Euclidean cluster extraction without the risk of segmenting a fallen person together with the adjacent scene elements. First of all, some false positive patches can be easily recognized, e.g. all the patches with less than 5 neighbouring positive patches in a radius of 0.5 \,m can be filtered out. Then, the negative patches are pushed aside, and the Euclidean clusters are extracted from the point cloud of the remaining positive patch centroids using a large distance threshold of 1.0\,m.

For each cluster, its OBB is calculated. Thus, depending on the OBB dimensions and the number of positive and negative patches in it, each cluster may be a fallen person or not. For each cluster, a feature vector $\mathbf{x_2}$ of size 9 has been devised. The complete list of features is given in Table \ref{features2}. In particular, the sample distances to the separating hyperplane returned by the former SVM turned out to be really useful. They have been exploited by means of an histogram with 4 bins for the distance intervals $[0, 0.25)$, $[0.25, 0.5)$, $[0.5, 1)$ and $[1, \infty)$. For each cluster, each histogram bin is filled with the positive patches whose distance to the hyperplane falls in the respective interval. Thus, the number of positive patches in each bin/interval gives 4 additional features. The whole feature vector is passed to a binary SVM classifier. After k-fold validation, a RBF kernel with the misclassification cost $C$ equal to 312.5 and the bandwidth $\gamma$ equal to \num{2.25e-3} turned out to be the best performing solution.
\begin{table}[h]
\caption{List of features calculated for each 3D cluster and their dimensionality.}
\label{features2}
\begin{center}
\begin{tabular}{c c}
\hline
Features & Dimensionality\\
\hline
OBB dimensions (width, height and depth) & 3\\
Number of positive patches & 1\\
Percentage of positive patches & 1\\
4-bin histogram of positive patch confidences & 4\\
\hline
Total number of features & 9\\
\hline
\end{tabular}
\end{center}
\end{table}
%


\subsection{Map Verification}
\label{subsec:map_verification}
\begin{figure}[!htb]
  \centering 
  \subfloat[\label{subfig:mapverificationa}]{%
    \includegraphics[width=.52\columnwidth ]{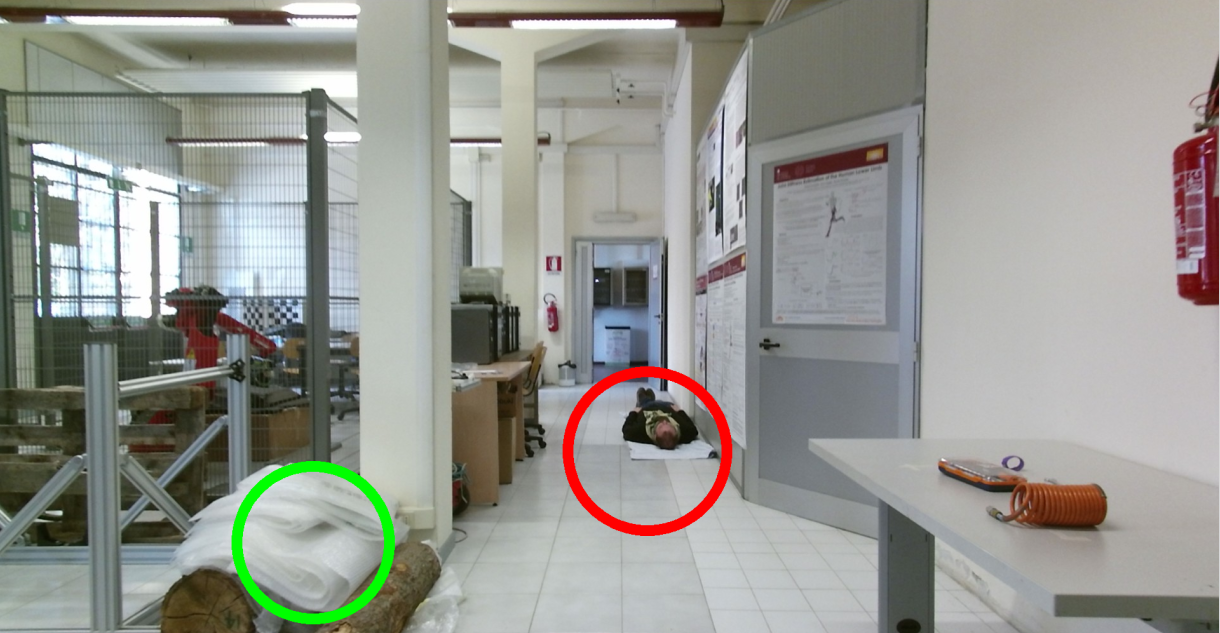}%
  }
  \subfloat[\label{subfig:mapverificationb}]{%
     \includegraphics[width=.48\columnwidth ]{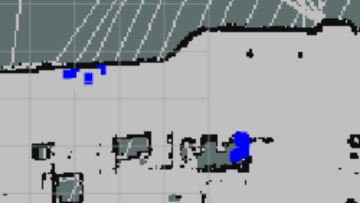}%
  }
  \caption{An example of successful false positive rejection performed by the map verification step: a) shows the furniture item raising some false positives, a tree trunk (in green) b) shows that these detections (the blue squares on the right) are located onto the map occupied space.}
  \label{fig:mapverification}
\end{figure}

A mobile robot navigates through the environment thanks to the information of two maps: a static one necessary to compute a collision-free plan with static objects, e.g. walls or furniture items, and a dynamic one necessary to avoid moving obstacles, e.g. people. In this work, the static map, which is usually acquired only once and for all, is exploited to implement a false positive rejection phase. Let the static map be defined as a set of cells $S = \{Cell_i, 0 \leq i \leq N\}$, where:
\begin{equation}
	Cell_i = \begin{cases}
       -1 & \textnormal{unknown content}\\
       0 & \textnormal{free space}\\
       0 < n \leq 1 & \textnormal{probability to be occupied}\\
     \end{cases}
\end{equation}
Thanks to the transformations computable with a 2D SLAM algorithm like \cite{grisetti2005improving,grisetti2007improved}, each single-view detection can be transformed from the camera coordinate system to the map coordinate system and projected to a cell map $Cell_i$. If the $Cell_i$ value is unknown ($-1$) or occupied by a static obstacle ($K \leq Cell_i \leq 1$ with $K = 0.30$), then the detection can be easily rejected. 
An example of successful false positive rejection is shown in Figure \ref{fig:mapverification}, in which a single-view detections falls on the static furniture, in this case a tree trunk. Indeed, given its geometric similarity to a lying person, the single-view algorithm may detect it as a person. The map verification allows to reject it, enhancing the final detection performances. This step handles also other challenging situations, like shelf glass surfaces which can be really noisy. 

\subsection{Merging Detections from Multiple Vantage Points}
\label{subsec:multiplevantagepoints}

The map is not the only robot feature that can enhance the detection performances. Indeed, in a typical scenario, the robot is patrolling a known environment. Thus, given that the location of each fallen person is mostly static, all the single-view detections available from the multiple points of view can be easily tracked. A detection may be a false positive from a certain view, while a true negative from many others. Moreover, the false positive detection rate is very low compared to the true positive one. Given these two facts, another contribution of this work is the exploitation of the detections available from the different vantage points.

After the map verification, the single-view detections are already expressed in the map reference system. In this section, an algorithm able to cluster or reject each of them is devised. Its output is a set $\mathfrak{P}$ of validated lying person locations $p_i$ in the map, formally $\mathfrak{P} = \{p_i, 0 \leq i \leq g\}$, where $g$ is the total number of people. Given each new detection $d = (loc, t)$, where $d.loc$ is its location in the map coordinate system and $d.t$ its timestamp, the set of clusters, formally $\mathfrak{C} = \{C_i: 0 \leq i \leq n\}$, is updated with the following rule:
\begin{equation}
C_i = \{d_j: ||d_j.loc - d_m.loc|| < \overline{th}, \forall j,m \in [0,k-1] \} \text{,}
\end{equation}
in which $\overline{th}$ is a user-defined threshold which indicates if a detection is close enough to be considered in the cluster or not, and $k$ the number of detections in the cluster. 


The set $\mathfrak{P}$ of fallen people is computed by a fixed-time periodic thread which analyses the set $\mathfrak{C}$. It updates the set $\mathfrak{P}$ by deleting the old detections and analysing the new ones in $\mathfrak{C}$. Indeed, in order to maintain a lightweight representation of $\mathfrak{C}$ and reject the false positives, whose frame rate is typically low, the old detections are discarded and a further check on the timestamp is performed. The pseudo-code of the whole procedure is reported in Algorithm \ref{alg:dofp_callback}, in which $\widehat{f}$ is the minimum detection frequency, $\widehat{t}$ is the maximum detection age and $\widehat{n}$ is the minimum number of detections in a cluster. Lines~\ref{alg:line:removal1}-\ref{alg:line:removal2} handle the time-based rejection on the basis of the maximum allowed age, while Lines~\ref{alg:line:classification1}-\ref{alg:line:classification2} reject the clusters whose detections have a low frame rate or are less than the minimum allowed.  

\begin{figure}[!htb]
  \centering
  \subfloat[\label{subfig:detection_multiframe}]{%
    \includegraphics[width=\columnwidth]{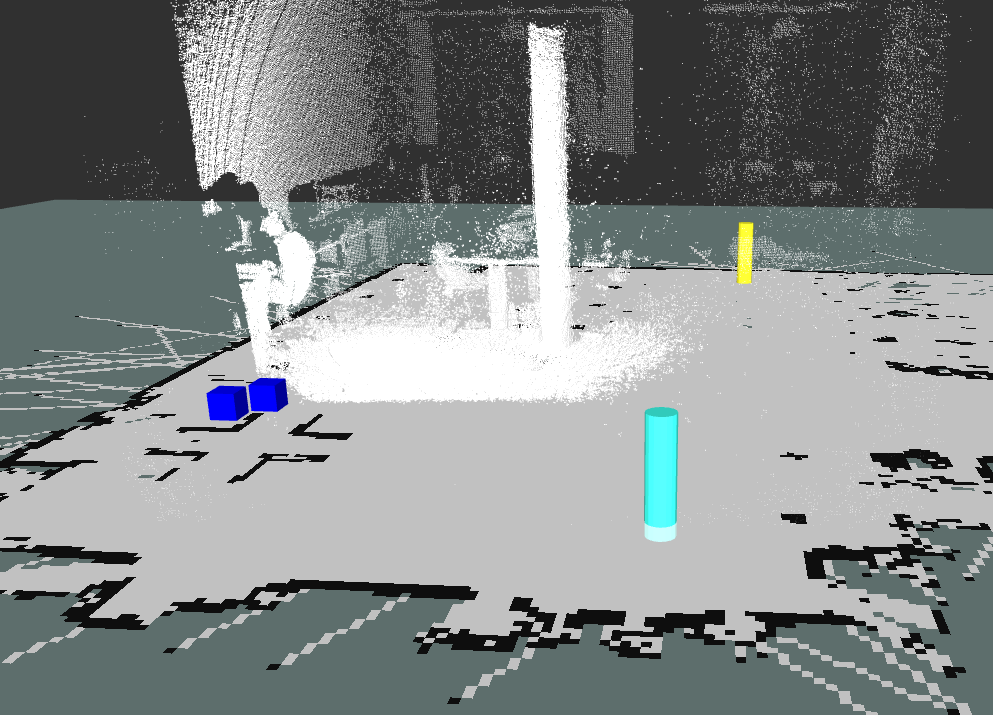}%
  }
  \\  
  \caption{The single-view detections projected on the 2D map are analysed by the \textit{multi-view analyser}. If they meet both the distance and time criteria, they are clustered. The white points compose the input point cloud, the blue cubes are the projected detections, here rejected false positives, and the coloured cylinders are the validated detection.}
  \label{fig:detection_multiframe}
\end{figure}

In our implementation, we used $\overline{th}$ equal to 1\,m, $\widehat{t}$ equal to 60\,s, $\widehat{f}$ equal to 1\,Hz and $\widehat{n}$ equal to 5. The use of the frame rate allows to set a low $\widehat{n}$, thus preventing over-fitting. The procedure is invoked by the periodic thread every 10 seconds. In Figure~\ref{fig:detection_multiframe}, the algorithm is shown in action.

\begin{algorithm}
\caption{Cluster validation for detecting fallen people exploiting multiple vantage points}\label{alg:dofp_callback}
\begin{algorithmic}[1]
\Procedure{VALIDATE\textunderscore CLUSTERS}{$\mathfrak{C}$, $\mathfrak{P}$, $\widehat{t}$, $\widehat{f}$, $\widehat{n}$}
\ForEach {$C_i \in \mathfrak{C}$} 
\For {$j \in [0,k - 2]$} \label{alg:line:removal1}
	\For {$o \in [j + 1, k - 1]$}
		\If {$|d_o.t - d_j.t| > \widehat{t}$}
		    \State {$index \gets $ARG\textunderscore MIN$(d_o.t, d_j.t)$}
			\State {$C_i \gets C_i \setminus d_{index}$}
		\EndIf
	\EndFor
\EndFor \label{alg:line:removal2}
\State {$t_m \gets \min_{d \in C_i}\{d.t\}$} \label{alg:line:classification1}
\State {$t_M \gets \max_{d \in C_i}\{d.t\}$}
\State {$f_i \gets $ $\frac{||\mathfrak{C}||}{t_M - t_m}$}
\If {$f_i \geq \widehat{f} \text{ and } ||C_i|| \geq \widehat{n}$}
\State {$loc_i \gets \sum_{d \in C_i} \frac{d.loc}{||C_i||}$}
\State {$\mathfrak{P} \gets \mathfrak{P} \cup loc_i$}
\State {$\mathfrak{C} \gets \mathfrak{C} \setminus C_i$} \label{alg:line:classification2}
\EndIf
\EndFor
\State {\Return $\mathfrak{C}$, $\mathfrak{P}$}
\EndProcedure
\end{algorithmic}
\end{algorithm}

\section{RESULTS}
\label{results}

The detection of fallen people is a challenging problem also because of the lack of public datasets. For this reason, another contribution of this work is the release of the IASLAB-RGBD Fallen Person Dataset\footnote{\url{http://robotics.dei.unipd.it/117-fall}}. On it, 4 common metrics, the detection $accuracy$, $precision$, $recall$ and $F_{0.5}$ score, are evaluated for each presented method. If $TP$, $TN$, $FP$ and $FN$ are the true positives, true negatives, false positives and false negatives, then these metrics are defined as in the following:
\begin{equation}
accuracy = \frac{TP + TN}{TP + FP + TN + FN}
\end{equation}
\begin{equation}
precision = \frac{TP}{TP + FP}
\end{equation}
\begin{equation}
recall = \frac{TP}{TP + FN}
\end{equation}
\begin{equation}
F_{0.5} = \frac{(1 + 0.5^2)*precision*recall}{0.5^2*precision + recall} \text{,}
\end{equation}
where the $F_{0.5}$ score, already proposed in \cite{volkhardt2013fallen}, is an harmonic average of precision and recall promoting an high precision, i.e. a low number of false positives. In addition, given the impossibility to compare with other existent and similar approaches, the baseline to which our algorithms are compared is a simple approach based on the Euclidean cluster extraction. This way, it will be clear how important the use of patches is in order to handle cluttered scenes. Finally, a detailed analysis of the running times is provided.

\subsection{IASLAB-RGBD Fallen Person Dataset}

This dataset consists of several RGB-D frame sequences containing 15 different people. It has been acquired in two different laboratory environments, the \textit{Lab A} and the \textit{Lab B}, by means of a Microsoft Kinect One V2, placed on a pedestal or on our mobile robot. The \textit{Lab A} is bigger and useful to test whether the algorithm can find people in the full distance range of the sensor (up to 5\,m). The \textit{Lab B} is smaller and more similar to a real domestic scenario. It is more cluttered and contains a sofa. It comprehends also glass surfaces which can be very noisy. For the sake of explanation, the dataset can be divided into three parts:
\begin{enumerate}
\item Part 1 includes 360 RGB-D frames acquired from 3 static pedestals. It is composed of several views of 10 people, which have been asked to lie in 12 different poses, 6 from the back and 6 from the front. Each person has been manually segmented in 3D;
\item Part 2 includes 4 sequences of RGB-D frames, for a total of 15932 frames, acquired from a mobile robot during its patrolling task in the \textit{Lab A}. People lie in 4 different fixed locations;
\item Part 3 includes 4 sequences of RGB-D frames, for a total of 9391 frames, acquired from a mobile robot during its patrolling task in the \textit{Lab B}. People lie in 4 different fixed locations.
\end{enumerate}
Training and test splits are also available. Some images of the dataset will be reported when discussing the results even if our approach does not exploit the RGB info.

The first classifier of the \textit{single-view detector} has been trained on thousand of patches extracted from the frames in Part 1 and Part 2 and tested on patches extracted from the frames in Part 1 and 3. All the positive samples have been taken from Part 1. The 70-30 train-test split of the segmented fallen people in Part 1 is also available. Negative samples have been taken from the \textit{Lab A} (just 24 frames out of 15932), the \textit{Lab B} (just 32 frames out of 9391) and the NYU Depth Dataset V2~\cite{Silberman2012indoor} (just 35 out of 1449), which contains thousands of indoor scenes for scene understanding. Only some of the negative samples have been used for balancing the number of positive and negative samples. 

The second classifier of the \textit{single-view detector} has been trained on clusters extracted from the frames in Part 2 (\textit{Lab A}) and tested on clusters extracted from the frames in Part 3 (\textit{Lab B}). Approximately, for the training, the 15\% of all the available frames has been considered.

Not only the \textit{single-view detector} but also the \textit{multi-view analyser} has been tested on Part 3. Indeed, both Part 2 and 3 comprehend the entire robot transformation tree. Given that the position of the fallen people in the 2D map is known, this allows to calculate the performance indices automatically by checking if the location of the detected cluster centroid is close (at a distance less or equal to 1\,m) to the ground truth centroid of a person position in the 2D map.


\subsection{Validation}


The presented methods have been quantitatively evaluated on the IASLAB-RGBD Fallen Person Dataset. First, the separated evaluation of each classifier is presented. Then, the entire pipeline has been evaluated on both rooms, the \textit{Lab A} and the \textit{Lab B}. As previously explained, both the classifiers have been trained on just a part of the frames in the \textit{Lab A} while they see the \textit{Lab B} for the first time. In particular, we present the results for each of the 3 contributions, the \textit{single-view detector} and the two modules of the \textit{multi-view analyser}: the map validation and the detection merging from multiple vantage points. Furthermore, given the impossibility to compare directly with \cite{volkhardt2013fallen}, the comparison baseline (B) is a simple approach not exploiting patches. It finds putative clusters by means of the Euclidean cluster extraction with a distance threshold of 0.10\,m, which is really low considering a voxel resolution of 0.06\,m and far less than the one required by our approach (1\,m).  The baseline classifies then each cluster on the basis of its position and its OBB size.

As previously mentioned, both classifiers of the \textit{single view detector} have been trained and tested on two different dataset splits. In both cases, K-fold validation with $K$ equal to 10 has been performed on the training set in order to find the optimal misclassification cost $C$ and bandwidth $\gamma$ values of the RBF kernel. As a preliminary evaluation, the SVM performances on the respective test sets are reported in Table~\ref{classifierperformances}.
\begin{table}[h]
\caption{Performances of the two classifiers on their test sets.}
\label{classifierperformances}
\begin{center}
\begin{tabular}{c c c c c}
\hline
Method & Accuracy & Precision & Recall & $F_{0.5}$\\
\hline
Classifier 1 (C1) & 0.89 & 0.93 & 0.84 & 0.91\\
Classifier 2 (C2) & 0.93 & 0.86 & 0.95 & 0.88\\
\hline
\end{tabular}
\end{center}
\end{table}

The results of the quantitative comparison between all the methods are shown in Table~\ref{methodcomparison1} and \ref{methodcomparison2}. Thanks to the patches, our methods outperform the baseline, not only in precision but also in recall. Furthermore, the map validation can further improve performances by rejecting some false positives. 
\begin{table}[h]
\caption{Performance comparison on the \textit{Lab A}.}
\label{methodcomparison1}
\begin{center}
\begin{tabular}{c c c c c}
\hline
Method & Accuracy & Precision & Recall & $F_{0.5}$\\
\hline
Baseline (B) & 0.88 & 0.65 & 0.33 & 0.54\\
Single-view (SV) & 0.90 & 0.77 & 0.78 & 0.77\\ 
SV + Map verification (MV) & 0.92 & 0.87 & 0.77 & 0.85 \\
\hline
\end{tabular}
\end{center}
\end{table}
\begin{table}[h]
\caption{Performance comparison on the \textit{Lab B}, never seen before by both classifiers.}
\label{methodcomparison2}
\begin{center}
\begin{tabular}{c c c c c}
\hline
Method & Accuracy & Precision & Recall & $F_{0.5}$\\
\hline
Baseline (B) & 0.84 & 0.64 & 0.26 & 0.50\\
Single-view  (SV) & 0.89 & 0.87 & 0.74 & 0.83\\ 
SV + Map verification (MV) & 0.90 & 0.92 & 0.72 & 0.87 \\
\hline
\end{tabular}
\end{center}
\end{table}

As shown in Table~\ref{multiviewperformance}, also the detection merging from multiple vantage points proved to be useful. It has been tested on each one of the eight frame sequences acquired in the two environments. Each time, even if the environment is the same, the navigation path can differ due to dynamic obstacles and the different positions of the lying people on the floor. After the 4 patrolling tasks of the \textit{Lab A}, each person is always detected and only once, a false positive is still present while, after the 4 patrolling tasks of the \textit{Lab B} (never seen before by both classifiers), each person is always detected and all the false positives are successfully rejected.
\begin{table}[h]
\caption{Performances of the \textit{multi-view analyser} on both environments. Each time, even if the environment is almost the same, the robot path can differ because of dynamic obstacles and different positions of the lying people on the floor.}
\label{multiviewperformance}
\begin{center}
\begin{tabular}{c c c c c}
\hline
Environment & TP/P & FP\\
\hline
Lab A (sequence 1) & 4/4 & 0 \\
Lab A (sequence 2) & 4/4 & 1 \\
Lab A (sequence 3) & 4/4 & 0 \\
Lab A (sequence 4) & 4/4 & 0 \\
Lab B (sequence 1) & 4/4 & 0 \\
Lab B (sequence 2) & 4/4 & 0 \\
Lab B (sequence 3) & 4/4 & 0 \\
Lab B (sequence 4) & 4/4 & 0 \\
\hline
\end{tabular}
\end{center}
\end{table}

\begin{figure*}[!htb]
  \centering
  \subfloat[\label{subfig:detection_multiframe_a}]{%
    \includegraphics[width=0.33\textwidth]{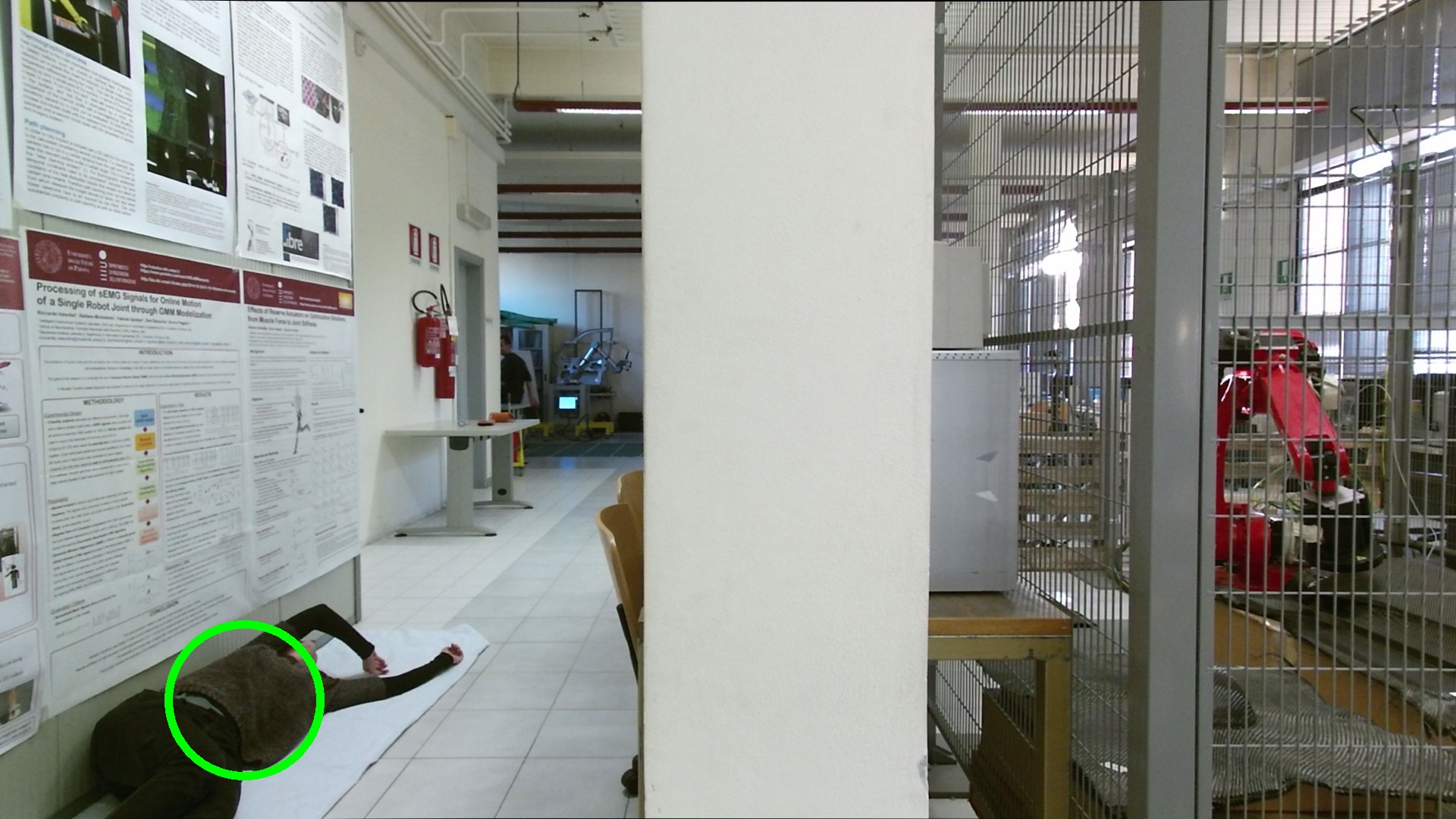}%
  }
  \subfloat[\label{subfig:detection_multiframe_b}]{%
     \includegraphics[width=0.33\textwidth]{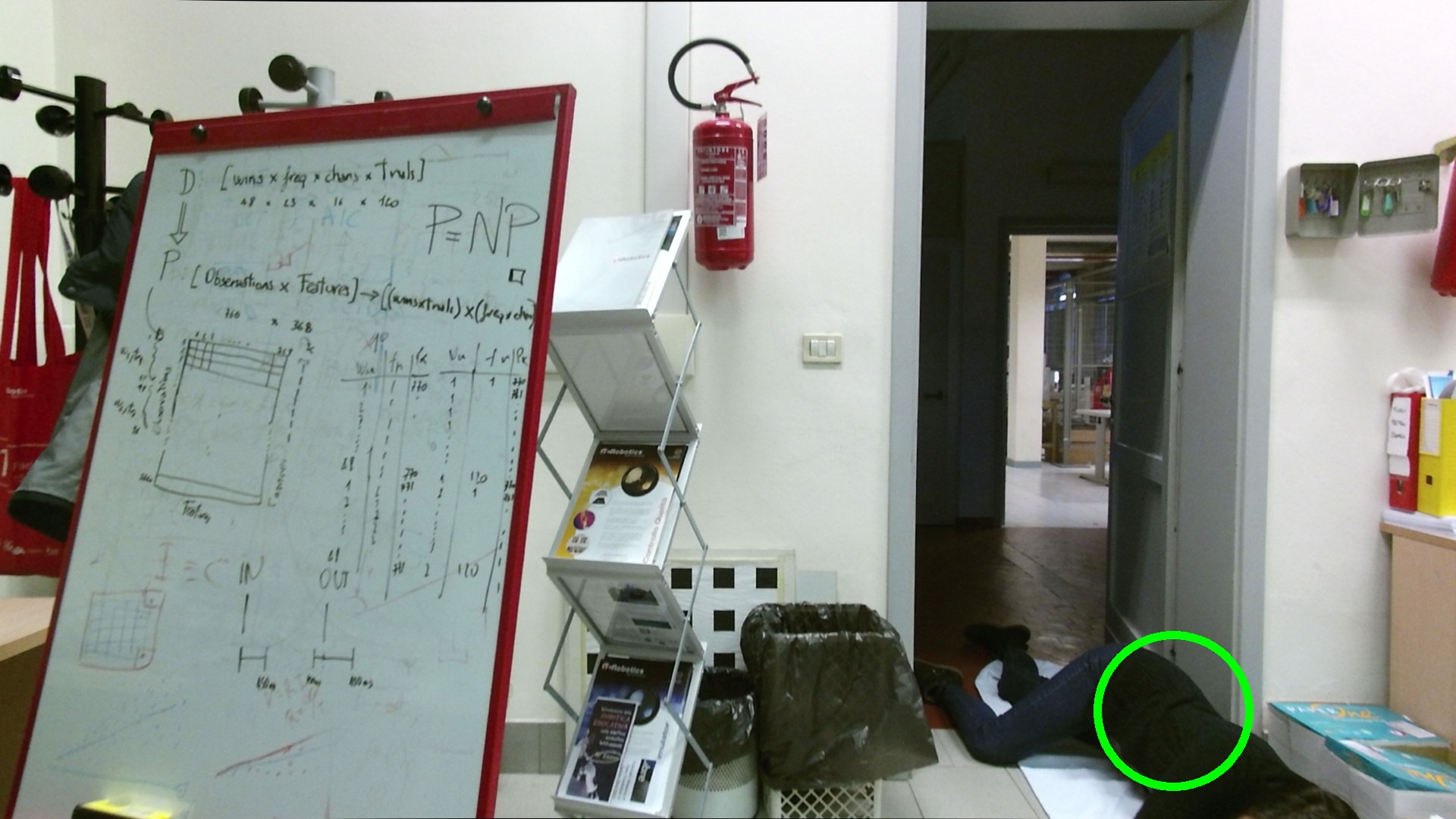}%
  }
  \subfloat[\label{subfig:detection_multiframe_b}]{%
     \includegraphics[width=0.33\textwidth]{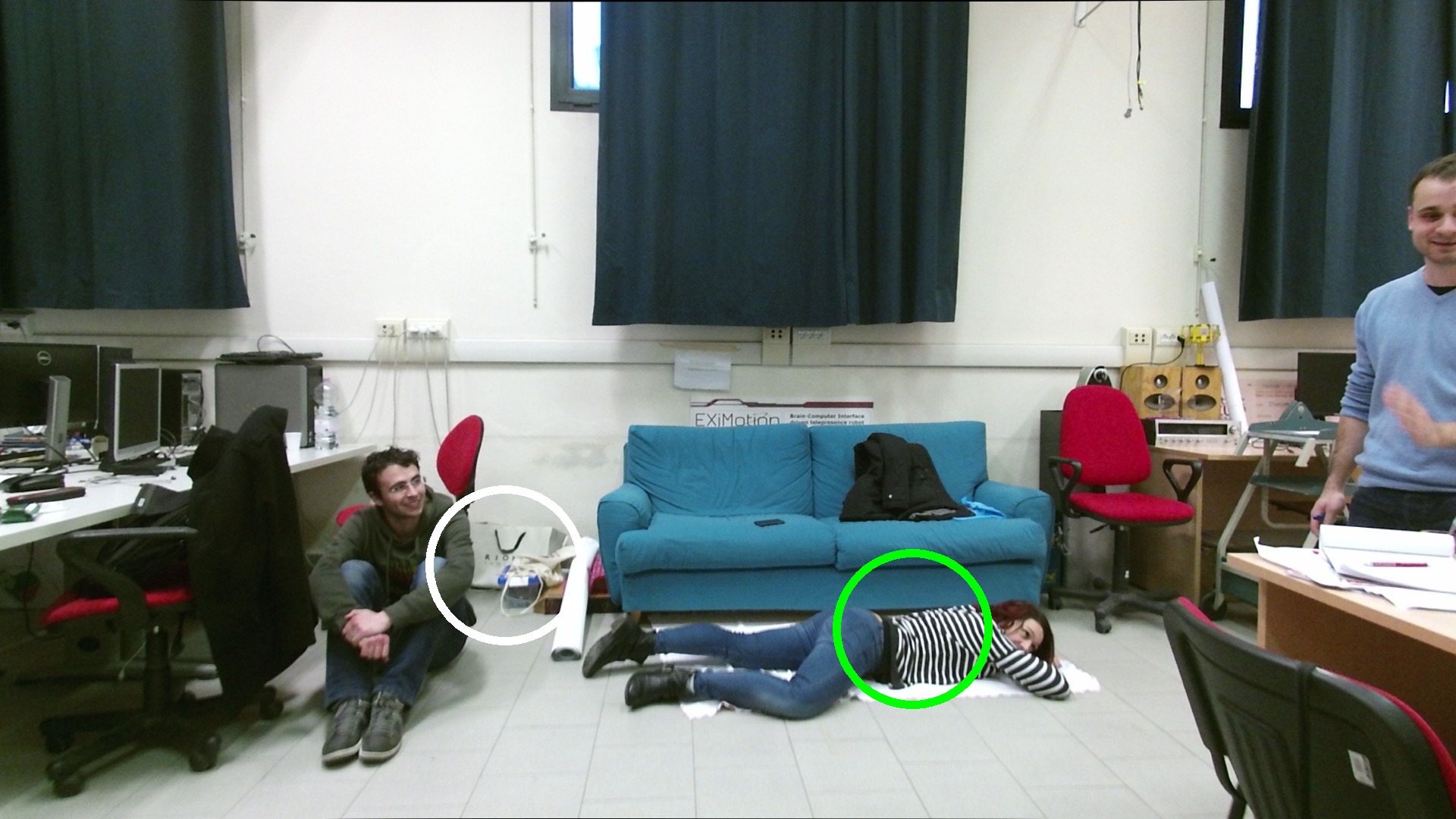}%
  }
  \\
  \subfloat[\label{subfig:detection_multiframe_a}]{%
    \includegraphics[width=0.33\textwidth]{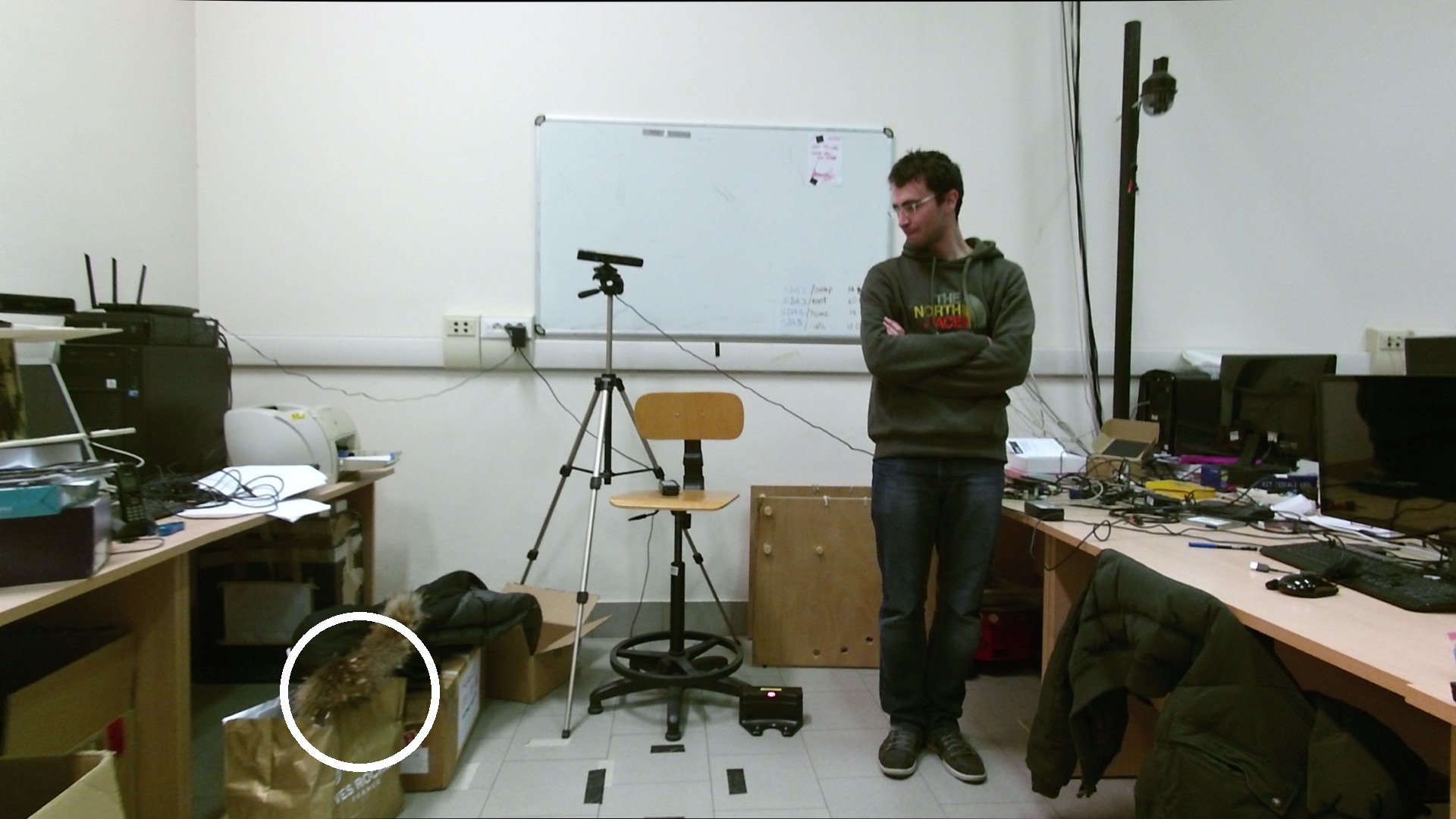}%
  }
  \subfloat[\label{subfig:detection_multiframe_b}]{%
     \includegraphics[width=0.33\textwidth]{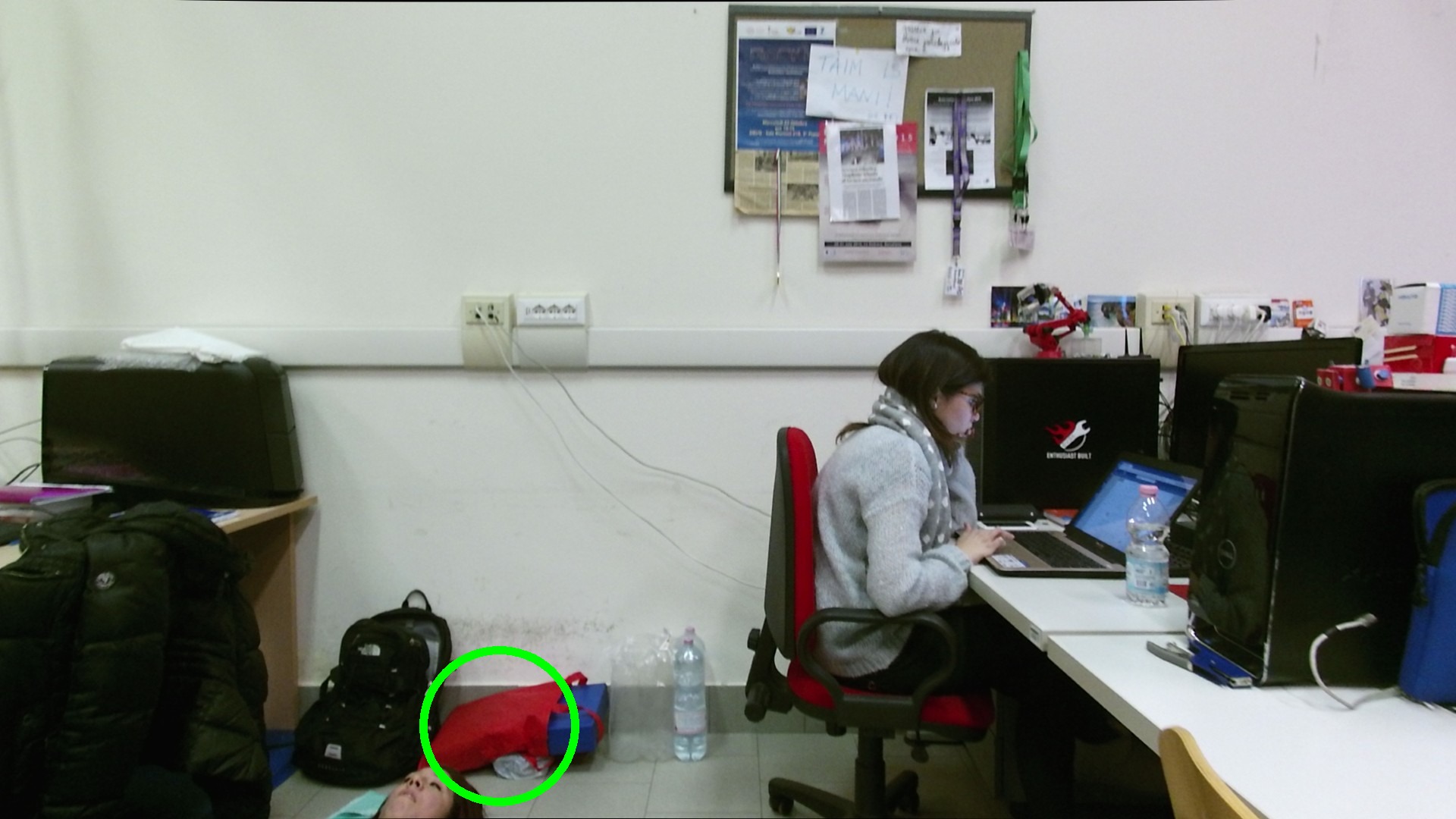}%
  }
  \subfloat[\label{subfig:detection_multiframe_b}]{%
     \includegraphics[width=0.33\textwidth]{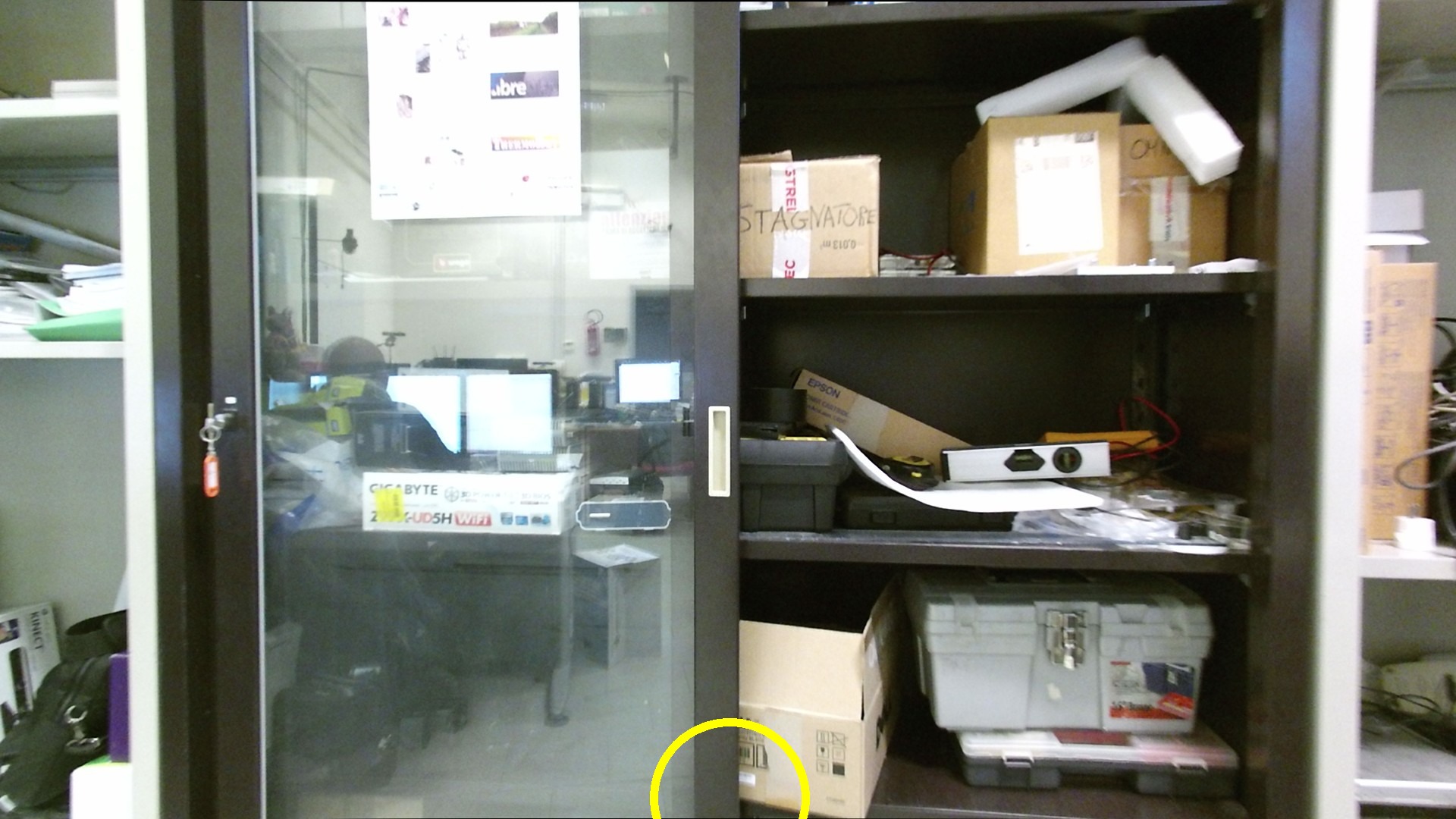}%
  }
  \\
  \subfloat[\label{subfig:detection_multiframe_a}]{%
     \includegraphics[width=0.44\textwidth]{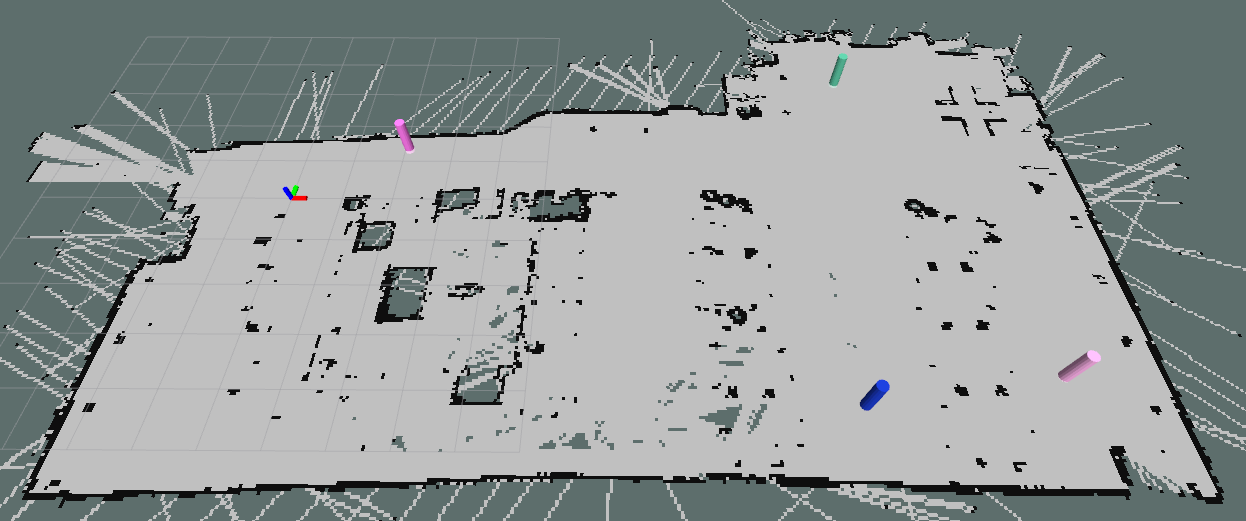}%
  }
  \subfloat[\label{subfig:detection_multiframe_b}]{%
     \includegraphics[width=0.33\textwidth]{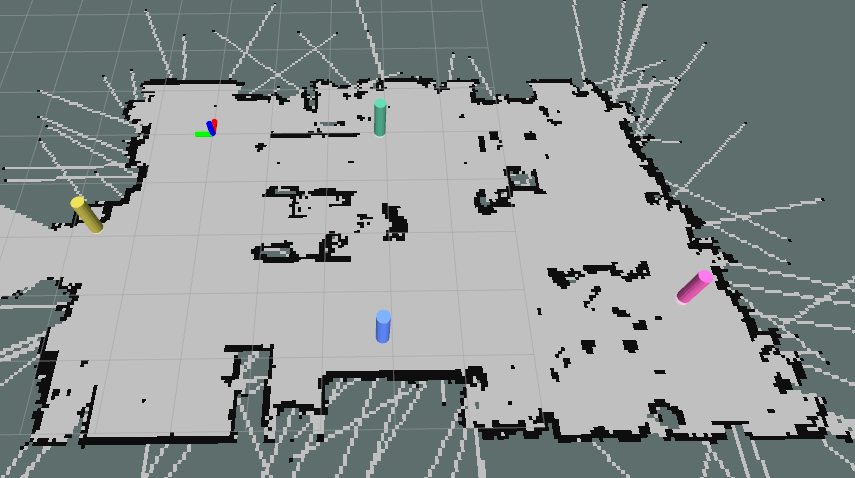}%
  }
  \caption{Qualitative results on the IASLAB-RGBD Fallen Person Dataset: (a)(b) even if the lying people can be very close to the wall or other scene elements, the \textit{single-view detector} can find them at a high detection rate; (c)(d) the \textit{single-view detector} can discard fake lying people, see the white circles; (e)(f) the \textit{single-view detector} may find some false positives in the presence of clutter (several close objects) or high noise (glass surfaces); (g)(e) the \textit{multi-view analyser} can reject both FP like in (e) thanks to the low frame rate or in (f) thanks to the map validation.}
  \label{fig:qualitative_results}
\end{figure*}

Finally, in Figure \ref{fig:qualitative_results}, some qualitative results are reported. They show the ability of the \textit{single-view detector} to find people in cluttered environments, see Figure \ref{fig:qualitative_results}(a)(b)(c)(d). Two difficult cases due to close objects or noisy regions, like glass surfaces, are also reported, see \ref{fig:qualitative_results}(e)(f). Anyway, they are easily handled by the \textit{multi-view detector}, \ref{fig:qualitative_results}(g)(h).





\subsection{Runtime Analysis}

In Table~\ref{runtimes}, the running times of \textit{single-view detector} are reported. The algorithm is very efficient in terms of computing time proving to be an optimal choice for a mobile robot. Even if it is not yet fully parallelized, it can work in real-time at an average speed of 7.72\,fps. The test machine is a Dell Inspiron 15 7000 with an Intel Core i7-6700HQ CPU with 4 cores clocked at 2.60GHz, 16 GB of RAM and Linux Mint 17.3. Given that the \textit{multi-view analyser} is a daemon running in the background, its running times are of no interest and thereby not reported.
\begin{table}[h]
\caption{Average runtimes of the main steps of the proposed algorithm on our test machine (Intel Core i7-6700HQ CPU, 2.60GHz x 4).}
\label{runtimes}
\begin{center}
\begin{tabular}{c c}
\hline
Processing Stage & Runtime\\
\hline
Pre-processing and Oversegmentation & 10.27\,fps\\
Patch Feature Extraction & 105.98\,fps\\
SVM Classification 1 (per patch) & 0.84 \,$\mu$s \\
Cluster Feature Extraction & 2639.56\,fps\\
SVM Classification 2 (per cluster) & 0.04 \,$\mu$s \\
\hline
Total runtime & 7.72\,fps\\
\hline
\end{tabular}
\end{center}
\end{table}
%



\section{CONCLUSIONS}
\label{conclusions}

This paper presented a real-time and robust approach to detect fallen people lying on the floor in various positions and from different distances. A single-view algorithm, which draws upon recent developments in the semantic segmentation field and does not need restrictive distance thresholds to segment putative clusters, was fully integrated on a mobile robot. The map of the environment and the availability of many different vantage points allowed to reduce the number of false positives, further improving the final performances. The algorithms here presented were thoroughly validated on the IASLAB-RGBD Fallen Person Dataset, which was published online for the benefit of the research community. They clearly outperform a simple method based on a finer distance threshold. In the near future, we would like to validate not only the ability of the algorithm to detect, but also to semantically segment fallen people.
We also plan to extend the test bed with sequences taken from real apartments along with different navigation paths. Finally, it would be interesting to merge close similar patches before their classification in order to analyse bigger segments.

%






\section*{ACKNOWLEDGEMENT}

The research leading to these results has been partially supported by Omitech Srl. The authors would like to thank M. Munaro and S. Ghidoni for valuable discussions.

\bibliography{mybib}{}
\bibliographystyle{ieeetr}

\end{document}